\pdfoutput=1

\documentclass[11pt]{article}

\usepackage[]{acl} 

\usepackage{times}
\usepackage{latexsym}
\usepackage{epsfig}
\usepackage{graphicx}
\usepackage{amsmath}
\usepackage{amssymb}
\usepackage{bm}
\usepackage{textcomp}
\usepackage{enumitem}
\usepackage{xcolor}
\usepackage[T1]{fontenc}

\usepackage[utf8]{inputenc}

\usepackage{microtype}

\definecolor{mygreen}{rgb}{0.09, 0.45, 0.27}

\usepackage{multirow}
\usepackage{orcidlink}
\usepackage{pdfrender}

\usepackage{wrapfig}
\usepackage{subfig}
\usepackage{marvosym}
\usepackage{lipsum}
\usepackage{inconsolata}
\newcommand{\ie}{\textit{i}.\textit{e}.}
\newcommand{\eg}{\textit{e}.\textit{g}.}

\title{Improving Data Augmentation for Robust Visual Question Answering with Effective Curriculum Learning}

\author{Yuhang Zheng\thanks{Co-first authors with equal contributions.}\\
  Zhejiang University \\
  \texttt{itemzheng@zju.edu.cn} \\\And
  Zhen Wang$^\ast$ \\
  Zhejiang University \\
  \texttt{zju\_wangzhen@zju.edu.cn} \\\And
  Long Chen\thanks{Corresponding Author.} \\
  HKUST \\
  \texttt{longchen@ust.hk} \\
  }

\begin{document}
\maketitle

\begin{abstract}
Being widely used in learning unbiased visual question answering (VQA) models, Data Augmentation (DA) helps mitigate language biases by generating extra training samples beyond the original samples. While today's DA methods can generate robust samples, the augmented training set, significantly larger than the original dataset, often exhibits redundancy in terms of difficulty or content repetition, leading to inefficient model training and even compromising the model performance. To this end, we design an \textbf{E}ffective \textbf{C}urriculum \textbf{L}earning strategy \textbf{ECL} to enhance DA-based VQA methods. Intuitively, ECL trains VQA models on relatively ``easy'' samples first, and then gradually changes to ``harder'' samples, and less-valuable samples are dynamically removed. Compared to training on the entire augmented dataset, our ECL strategy can further enhance VQA models' performance with fewer training samples. Extensive ablations have demonstrated the effectiveness of ECL on various methods.
\end{abstract}

\section{Introduction}

Visual Question Answering (VQA) --- answering natural language questions about the given visual content --- has raised unprecedented attention due to its multi-modal nature. However, today's VQA models still suffer from severe \textbf{language biases}~\citep{agrawal2018don}, over-relying on linguistic correlations rather than multi-modal reasoning. To realize robust VQA, recent works~\citep{chen2020counterfactual,chen2021counterfactual,kolling2022efficient,agarwal2020towards,gokhale2020mutant,gokhale2020vqa,boukhers2022coin,tang2020semantic,kant2021contrast,bitton2021automatic,narjes2022inductive,wang2021cross} employ various data augmentation (DA) techniques by generating extra training samples, to enhance VQA models' performance on both \emph{in-domain} (ID)~\citep{goyal2017making} and \emph{out-of-distribution} (OOD) datasets~\citep{agrawal2018don}.

Early synthetic-based DA methods synthesizing new samples through visual region/word editing or regeneration~\citep{chen2020counterfactual,chen2021counterfactual,kolling2022efficient,agarwal2020towards}. However, these synthetic samples are error-prone~\citep{tang2020semantic} and require extra human annotations for reasonable answers~\citep{gokhale2020mutant}. Later, SimpleAug~\citep{kil2021discovering} avoids synthetic samples by randomly pairing images and questions with pseudo ground-truth answers. Nonetheless, its heuristic rule design leads to inaccuracies, limited coverage, and dependence on human annotations. Recent state-of-the-art DA method KDDAug~\citep{chen2022rethinking} further relaxes the requirement for VQ pairs automatically, which generates more robust pseudo answers based on knowledge distillation. It achieves superior performance on both ID and OOD settings with fewer samples, and avoids all the mentioned weaknesses in existing DA methods.

Subsequently, after data augmentation, we can obtain an augmented training set several times the original dataset's size. However, these augmented samples may exhibit repetitive difficulty levels or content, posing challenges to training efficiency and potentially compromising model performance. While several pioneering works~\cite{lao2021superficial,pan2022causal} tried to apply the curriculum learning (CL)~\cite{bengio2009curriculum} for VQA models, which first trains the model on easy samples and gradually extends to hard samples, these CL-based methods mainly focusing on debiasing the original data instead of training the augmented ones. 

To this end, thanks to the robust samples generated by KDDAug~\citep{chen2022rethinking}, we design an \textbf{E}ffective \textbf{C}urriculum \textbf{L}earning (CL) based training strategy \textbf{ECL}, helping the VQA models focus on learning the unbiased augmented data without caring about debiasing. During each training iteration, we calculate a difficult score for every augmented sample dynamically based on the current state-of-the-art VQA model. Following the spirits of CL, we begin the model's training with easier samples and gradually transfer to harder ones according to their difficult scores. Meanwhile, to further boost the training process, we propose to remove \emph{less-valuable} samples during the CL training stage. For more comprehensive evaluations, we evaluated KDDAug with our ECL (denoted as KDDAug-ECL) on two challenging benchmarks: VQA v2 and VQA-CP. Moreover, we boost multiple different VQA architectures to achieve state-of-the-art performance by applying KDDAug-ECL. Compared with training on the entire augmented dataset, our ECL strategy can further help VQA models benefit from augmented data and achieve better performance with fewer training samples.

\section{Related Work} 

\noindent\textbf{Data Augmentation in VQA.} To address language biases in VQA, DA serves as a model-agnostic debiasing method by generating augmented training samples. Apart from mainstream synthetic-based methods, alternative DA approaches include generating negative samples through random selection of images or questions~\cite{teney2020value,zhu2020overcoming} and composing reasonable image-question (VQ) pairs as positive training samples~\cite{kil2021discovering}. However, they often exhibit either a significant ID performance drop~\cite{chen2020counterfactual,chen2021counterfactual,teney2020value,kolling2022efficient} or rely on human annotations for answer assignment, lacking generality~\cite{kafle2017data,gokhale2020mutant,gokhale2020vqa,kil2021discovering,narjes2022inductive}. In contrast, KDDAug~\cite{chen2022rethinking} effectively mitigates these challenges, achieving the best trade-off results on both ID and OOD settings.

\noindent\textbf{Curriculum Learning (CL).}  The easy-to-hard spirit of CL~\cite{bengio2009curriculum} is inspired by the learning process of us humans. This human-like effective training paradigm has been extensively exploited in different vision-language tasks~\cite{dong2021dual,seo2020reinforcing,zheng2022dual,yao2021visual}.
Recent works~\cite{lao2021superficial,pan2022causal} tried to introduce CL into VQA to reduce language biases by helping VQA models gradually focus on more biased samples~\cite{lao2021superficial} or gradually increasing the importance of visual features in the training phase~\cite{pan2022causal}. Different from complex CL of existing debiasing methods focusing on the training of original data, we design a simple yet effective CL training strategy that lets the model focus on the learning of augmented data without caring about debiasing.

\begin{figure}[t]
    \centering
    \includegraphics[width=0.99\linewidth]{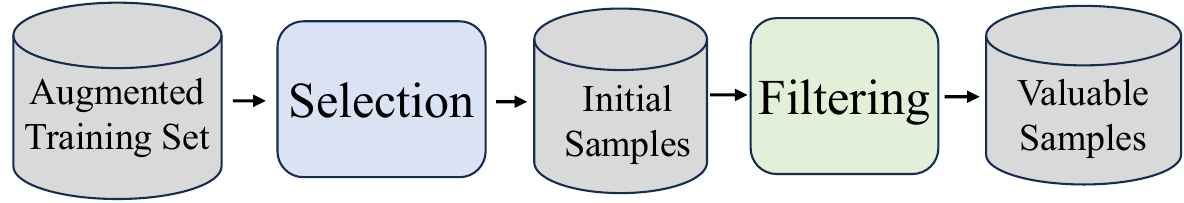}
    \vspace{-1em}
    \caption{The pipeline of our training strategy.}
    \vspace{-1em}
    \label{fig:pipeline}
\end{figure}


\begin{table*}[t]
	\begin{center}
	    \scalebox{0.73}{
	        \begin{tabular}{| l | l | c c c c | c c c c| c | }
			\hline
			\multirow{2}{*}{\textbf{Base}} & \multirow{2}{*}{\textbf{Models}} & \multicolumn{4}{c|}{VQA-CP v2 test} & \multicolumn{4}{c|}{VQA v2 val} &  \multirow{2}{*}{\textbf{HM}}  \\
			& & \textbf{All} & Y/N & Num & Other & \textbf{All} & Y/N & Num & Other &  \\
		    \hline
		        \multirow{3}{*}{UpDn~\cite{anderson2018bottom}} & Baseline$^*$ & 39.85 & 42.66 & 12.18 & 45.98 & \textbf{63.30} & \textbf{81.06} & \textbf{42.46} & \textbf{55.32} & 48.91 \\
		        & KDDAug & 60.24 & 86.13 & 55.08 & 48.08 & 62.86 & 80.55 & 41.05 & 55.18 & 61.52 \\
                    & KDDAug-ECL & \textbf{61.80}$_{\textcolor{mygreen}{\textbf{+1.56}}}$ & \textbf{89.29} & \textbf{57.17} & \textbf{48.67} & 63.20$_{\textcolor{mygreen}{\textbf{+0.34}}}$ & 80.97 & 41.50 & 55.42 & \textbf{62.49}$_{\textcolor{mygreen}{\textbf{+0.97}}}$ \\
            \hline
                \multirow{3}{*}{LMH~\cite{clark2019don}} & Baseline$^*$ & 53.87 & 73.31 & 44.23 & 46.33 & 61.28 & 76.58 & \textbf{55.11} & 40.69  & 57.24 \\
		        & KDDAug & 59.54 & 86.09 & 54.84 & 46.92 & 62.09 & 79.26 & 40.11 & {54.85}  & 60.79 \\
                & KDDAug-ECL & \textbf{60.39}$_{\textcolor{mygreen}{\textbf{+0.85}}}$ & \textbf{87.79} & \textbf{56.17} & \textbf{47.19} & \textbf{62.91}$_{\textcolor{mygreen}{\textbf{+0.82}}}$ & \textbf{80.66} & 41.38 & \textbf{55.10} &  \textbf{61.62}$_{\textcolor{mygreen}{\textbf{+0.83}}}$ \\
            \hline
                \multirow{3}{*}{RUBi~\cite{cadene2019rubi}} & Baseline$^*$ & 46.84 & 70.05 & 44.29 & 11.85 & 52.83 & 54.74 & \textbf{41.56} & \textbf{54.38} & 49.66 \\
		        & KDDAug & {59.25} & {84.16} & {54.12} & {47.61} & {60.25} & {74.97} & 40.29 & 54.35 & 59.75 \\
              & KDDAug-ECL & \textbf{59.62}$_{\textcolor{mygreen}{\textbf{+0.37}}}$ & \textbf{84.37} & \textbf{55.89} & \textbf{47.68} & \textbf{60.30}$_{\textcolor{mygreen}{\textbf{+0.05}}}$ & \textbf{75.56} & 40.18 & 54.03 & \textbf{59.96}$_{\textcolor{mygreen}{\textbf{+0.21}}}$ \\
		\hline
                \multirow{3}{*}{CSS$^+$~\cite{chen2021counterfactual}} & Baseline$^*$ & 59.19 & 83.54 & 51.29 & 48.59 & 58.91 & 71.02 & 39.76 & 54.77 & 59.05 \\
		        & KDDAug & 61.14 & 88.31 & 56.10 & 48.28 & 62.17 & 79.50 & \textbf{40.57} & 54.71 & {61.65} \\
                & KDDAug-ECL & \textbf{62.06}$_{\textcolor{mygreen}{\textbf{+0.92}}}$ & \textbf{89.57} & \textbf{56.74} & \textbf{49.10} & \textbf{62.43}$_{\textcolor{mygreen}{\textbf{+0.26}}}$ & \textbf{80.03} & 40.49 & \textbf{54.86} & \textbf{62.24}$_{\textcolor{mygreen}{\textbf{+0.59}}}$ \\
			\hline
    		\end{tabular}
	    }
	\end{center}
        \vspace{-1em}
	\caption[]{Accuracies (\%) of different VQA architectures. $^*$ indicates our reimplementation using official codes.}
        \vspace{-1em}
    \label{tab:architecture_agnostic}
\end{table*}

\section{Effective Curriculum Learning (ECL)}

\noindent\textbf{DA-based VQA baseline.} Given the original training set $\mathcal{D}_{\text{orig}}$ and augmented training set $\mathcal{D}_{\text{aug}}$ generated by KDDAug~\citep{chen2022rethinking}, VQA models are first pre-trained with only the original training set $\mathcal{D}_{\text{orig}}$, then fine-tuned with augmented training set $\mathcal{D}_{\text{aug}}$. In order to more effectively train VQA models on the augmented samples, we introduce the easy-to-hard idea of curriculum learning~(CL). Benefiting from the robust samples generated by KDDAug, we can pay more attention to the training process instead of debiasing when designing CL. As shown in Figure~\ref{fig:pipeline}, ECL consists of two key steps: the difficulty-based sample selection and the value-based sample filtering.

\noindent\textbf{Difficulty-based Sample Selection.}
\emph{1) Difficulty Measurement.} A typical CL framework consists of two components: difficulty measurer and training scheduler, respectively~\cite{wang2021survey}. 
In fact, there is no uniform standard to judge the difficulty of a VQA sample. A feasible solution is utilizing the loss value to measure the difficulty of the sample for the current model. For two augmented samples, the sample with the smaller loss value means that the prediction of the VQA model is closer to the assigned ground-truth answer, \ie, it can be considered that this sample is less difficult for the current VQA model. 
To this end, we employ self-paced learning~(SPL)~\cite{kumar2010self}, which can control the training process and optimize the VQA model jointly in a unified framework. Specifically, we introduce a binary weight $v_i \in \{0, 1\}$ for each augmented training sample, and re-formulate the loss function as:
\begin{equation}
\small
L_{SPL} = \textstyle{\sum}_{i=1}^{|\mathcal{D}_{aug}|} \big( v_i \cdot \texttt{XE}( pred_i, a^t_i) - \lambda v_i \big),
\label{eq:spl}
\end{equation}
where \texttt{XE} is the cross-entropy loss function, $pred_i$ and $a^t_i$ are $i$-th augmented sample's prediction and answer, respectively. $\lambda v_i$ is a self-paced regularizer that controls the training process by $\lambda$.

\noindent\emph{2) Weight Optimization.} Following SPL, we alternately optimize the VQA model and the weights $v_i$ of augmented samples. Concretely, we first optimize sample weights $v_i$ while keeping the VQA model frozen. By observing the Eq.~(\ref{eq:spl}), we can find that to minimize $L_{SPL}$, we only need to set $v_i = 0$ for those samples whose loss values are larger than $\lambda$, \ie, $\texttt{XE}(pred_i, a_i^t) - \lambda > 0$, which means these samples do not participate in training. Then, with fixed $v_i$, we optimize the VQA model. Obviously, by gradually increasing the value of $\lambda$, the VQA model is first trained on easier samples and then gradually extends to harder samples.

\noindent\emph{3) Sample Selection.} We adjust the value of $\lambda$ to make the size of training samples grow linearly. Specifically, before the training of each epoch, we utilize the current VQA model to calculate the loss value for each augmented sample, and rank them in ascending order. We take the loss value of the $P$-th sample as $\lambda$, where $P$ can be calculated by:
\begin{equation}
\small
    P = \big\lceil min(p * t, 1) * |D_{aug}| \big\rceil, 
\end{equation}
where $p \in [0, 1] $ is a hyperparameter to control the growth progress of the size of training samples and $t$ is the current epoch. Samples with loss values smaller than $\lambda$ are selected as the \emph{Initial Samples}.

\noindent\textbf{Value-based Sample Filtering.}
\emph{1) Value Measurement.} Considering that samples that can be easily answered by the VQA model in the early stage may have a relatively limited impact to the model at the later training stages, which even may cause the issue of knowledge redundancy, we design a more effective CL training strategy by dynamically removing these less-valuable samples. Following~\cite{li2022knowledge}, we measure the value of augmented samples towards current VQA model by its \textbf{prediction entropy} based on two verified insights: 1) The average prediction entropy loss decreases drastically as CL proceeds~\cite{rafael2019when}. 2) The informative knowledge~(\ie, more valuable samples) tends to require more training time to fit well~\cite{shen2021is}. Thus we obtain $i$-th augmented sample's value by:
\begin{equation}
\small
    V_i = H(pred_i) = \textstyle{\sum}_{a \in \mathcal{A}} -pred_i[a]* \log pred_i[a],
\end{equation}
$pred[a]$ is the predicted probability for answer $a$. 

\noindent\emph{2) Sample Filtering.} After getting the value of the augmented samples, we remove samples whose value is less than the threshold $\gamma$. Similar to $\lambda$, we regard the $Q$-th smallest value as the threshold $\gamma$, where $Q$ increases linearly with epoch $t$:
\begin{equation}
\small
    Q = \big\lceil min(q * t, 1) * |D_{aug}| \big\rceil, 
\end{equation}
$q \in [0, 1] $ is a hyperparameter to control the pace. Samples with values larger than $\gamma$ are then kept as the \emph{Valuable Samples} for model training. 

\noindent\textbf{Advantages.} Compared with existing CL, our designed ECL has two advantages: 1) ECL automatically determines the difficulty of samples and learning progress according to the current state of VQA model, which avoids complicated course design. 2) ECL removes less-valuable samples in time, which makes CL more efficient and effective.

\begin{table*}[t]
	\begin{center}
	    \scalebox{0.75}{
    		\begin{tabular}{| l | c c c c | c c c c| c |}
			\hline
    			\multirow{2}{*}{\textbf{Models}} & \multicolumn{4}{c|}{VQA-CP v2 test} & \multicolumn{4}{c|}{VQA v2 val} &  \multirow{2}{*}{\textbf{HM}} \\
	    		& \textbf{All} & Y/N & Num & Other & \textbf{All} & Y/N & Num & Other &  \\
		    \hline
		    	UpDn~\cite{anderson2018bottom}$_{\texttt{CVPR'18}}$ & 39.74 & 42.27 & 11.93 & 46.05 & 63.48 & 81.18 & 42.14 & 55.66 & 48.88\\
			 	~~+CVL~\cite{abbasnejad2020counterfactual}$_{\texttt{CVPR'20}}$  & 42.12 & 45.72 & 12.45 & 48.34 &  --- & --- & ---  & --- & --- \\
			 
			 	~~+Unshuffling~\cite{teney2020unshuffling}$_{\texttt{ICCV'21}}$  & 42.39 & 47.72 & 14.43 & 47.24 & 61.08 & 78.32 & 42.16 & 52.81 & 50.05 \\
			 
			    ~~+CSS~\cite{chen2020counterfactual}$_{\texttt{CVPR'20}}$  & 41.16 & 43.96 & 12.78 & 47.48 & --- & --- & --- & --- & --- \\ 
                ~~+MUTANT$^\dagger$~\cite{gokhale2020mutant}$_{\texttt{EMNLP'20}}$ &  50.16 & 61.45 & 35.87 & \textbf{50.14} & --- & --- & --- & --- & --- \\
                
                ~~+SimpleAug~\cite{kil2021discovering}$_{\texttt{EMNLP'21}}$ &  52.65 & 66.40 & 43.43 & 47.98 & \textbf{64.34} & \textbf{81.97} & \textbf{43.91} & \textbf{56.35} & 57.91 \\
                ~~+CSS$^+$~\cite{chen2021counterfactual}$_{\texttt{TPAMI'23}}$  & 40.84 & 43.09 & 12.74 & 47.37 & --- & --- & --- & --- & --- \\
                ~~+\textbf{KDDAug}~\cite{chen2022rethinking}$_{\texttt{ECCV'22}}$ &  60.24 & 86.13 & 55.08 & 48.08 & 62.86 & 80.55 & 41.05 & 55.18 & 61.52 \\

                ~~+\textbf{KDDAug-ECL} & \textbf{61.80} & \textbf{89.29} & \textbf{57.17} & 48.67 & 63.20 & 80.97 & 41.50 & 55.42 & \textbf{62.49} \\
            \hline
            	LMH$^*$~\cite{clark2019don}$_{\texttt{EMNLP'19}}$ & 53.87 & 73.31 & 44.23 & 46.33 & 61.28 & 76.58 & 55.11 & 40.69 & 57.24 \\

                ~~+CSS~\cite{chen2020counterfactual}$_{\texttt{CVPR'20}}$ &  58.95 & 84.37 & 49.42 & 48.21 & 59.91 & 73.25 & 39.77 & 55.11 & 59.43 \\
						
			    ~~+SimpleAug~\cite{kil2021discovering}$_{\texttt{EMNLP'21}}$ &  53.70 & 74.79 & 34.32 & 47.97 & {62.63} & 79.31 & \textbf{41.71} & \textbf{55.48} & 57.82 \\
			
			    ~~+ECD~\cite{kolling2022efficient}$_{\texttt{WACV'22}}$ &  59.92 & 83.23 & 52.29 & \textbf{49.71} & 57.38 & 69.06 & 35.74 & 54.25 & 58.62 \\

			    ~~+CSS$^+$~\cite{chen2021counterfactual}$_{\texttt{TPAMI'23}}$  & 59.54 & 83.37  & 52.57  & 48.97  & 59.96 & 73.69  & 40.18  & 54.77  & 59.75 \\

			    ~~+\textbf{KDDAug}~\cite{chen2022rethinking}$_{\texttt{ECCV'22}}$ &  59.54 & 86.09 & 54.84 & 46.92 & 62.09 & 79.26 & 40.11 & 54.85 & 60.79 \\
                    ~~+\textbf{KDDAug-ECL} &  60.39 & 87.79 & 56.17 & 47.19 & \textbf{62.91} & \textbf{80.66} & 41.38 & 55.10 & 61.62\\
			
			    ~~+\textbf{CSS$^+$}+\textbf{KDDAug} &  61.14 & 88.31 & 56.10 & 48.28 & 62.17 & 79.50 & 40.57 & 54.71 & 61.65 \\ 

                   ~~+\textbf{CSS$^+$}+\textbf{KDDAug-ECL} & \textbf{62.06} & \textbf{89.57} & \textbf{56.74} & 49.10 & 62.43 & {80.03} & 40.49 & 54.86 & \textbf{62.24} \\ 
            \hline
    		\end{tabular}
	    }
	\end{center}
        \vspace{-1em}
        \caption[]{Accuracies (\%) on VQA-CP v2 and VQA v2 of SOTA models. $^*$ indicates the results from our reimplementation. ``MUTANT$^\dagger$" denotes MUTANT~\cite{gokhale2020mutant} only trained with \texttt{XE} loss.} 
	\label{tab:SOTA_v2}
\end{table*}

\begin{table}[t]
        \vspace{-1em}
	\begin{center}
        \scalebox{0.73}{
        	\begin{tabular}{|l|l| c c |c c | c |}
			\hline
			\multirow{2}{*}{Base} & \multirow{2}{*}{Models} & \multicolumn{2}{c|}{VQA-CP v2 test} & \multicolumn{2}{c|}{VQA v2 val} & \multirow{2}{*}{\textbf{HM}} \\
			& & \textbf{All} & Other & \textbf{All} & Other &  \\
		    \hline
		    \multirow{3}{*}{UpDn} 		        
                & KDDAug & 60.24 & 48.08 & 62.86 & 55.18 & 61.52 \\
                & ~~+CL & 61.75 & 48.55 & 62.93 & 55.00 & 62.33 \\
                & ~~+ECL & \textbf{61.80} & \textbf{48.67} & 63.20 & 55.42 & \textbf{62.49}\\
            \hline
                \multirow{3}{*}{LMH} 
		    & KDDAug & 59.54 & 46.92 & 62.09 & {54.85} & 60.79 \\
                & ~~+CL & 60.39 & 47.13 & 62.80 & 54.92 & 61.57\\
                & ~~+ECL & \textbf{60.39} & \textbf{47.19} & \textbf{62.91} & \textbf{55.10} & \textbf{61.62}\\
            \hline
                \multirow{3}{*}{RUBi} 
		        & KDDAug & 59.25 & 47.61 & 60.25 & 54.35 & 59.75\\
                & ~~+CL & \textbf{59.97} & \textbf{47.73} & \textbf{60.64} & 54.20 & \textbf{60.30} \\
              & ~~+ECL & 59.62 & 47.68 & 60.30 & 54.03 & 59.96 \\
		\hline
                \multirow{3}{*}{CSS$^+$} 
		    & KDDAug & 61.14 & 48.28 & 62.17 & 54.71 & 61.65 \\
                & ~~+CL & 61.97 & 48.81 & 62.38 & 54.78 & 62.17 \\
                & ~~+ECL & \textbf{62.06} & \textbf{49.10} & \textbf{62.43} & \textbf{54.86} & \textbf{62.24}\\
			\hline
    		\end{tabular}
        }
	\end{center}
        \vspace{-1em}
	\caption[]{Accuracies of different VQA architectures. $^*$ indicates our reimplementation.}
	\label{tab:ECL}
\end{table}

\section{Experiments}

\subsection{Experimental Settings}

\noindent\textbf{Datasets and Metrics.} We evaluated ECL on two datasets: the ID benchmark \textbf{VQA v2}~\cite{goyal2017making} and OOD benchmark \textbf{VQA-CP v2}~\cite{agrawal2018don}. We followed standard VQA evaluation metric~\cite{antol2015vqa}, and used Harmonic Mean~(\textbf{HM}) of the accuracies on both two datasets (VQA v2 val \& VQA-CP test) to evaluate the trade-off between ID and OOD evaluations.

\noindent\textbf{Implementation Details.} We evaluated the effectiveness of ECL based on KDDAug~\citep{chen2022rethinking}. For comparison, we denote the models with training on augmented samples as \textbf{KDDAug}, and denote the models with training on both KDDAug and our ECL training strategy as \textbf{KDDAug-ECL}.

\subsection{Architecture Agnostic}

\noindent\textbf{Settings.} Since KDDAug is model-agnostic, we follow KDDAug and apply KDDAug-ECL to multiple different VQA models to validate the generalization of ECL, including \textbf{UpDn}, \textbf{LMH}, \textbf{RUBi}, and \textbf{CSS$^+$}. All the results are shown in Table~\ref{tab:architecture_agnostic}.

\noindent\textbf{Results.} ECL can further boost the improvements, pushing all models' performance to the state-of-the-art level. Particularly, the improvements are most significant in baseline UpDn (\eg, 0.97\% gains on HM). 
Furthermore, when KDDAug-ECL is applied to CSS$^+$, it can still improve the performance on both OOD and ID benchmarks, and achieve the best performance (\eg, 62.24\% on HM).

\subsection{Comparisons with State-of-the-Arts}

\noindent{\textbf{Settings.}} We incorporated the KDDAug-ECL into model \textbf{UpDn}, \textbf{LMH} and \textbf{CSS$^+$}, and compared them with the SOTA DA models both on VQA-CP v2 and VQA v2. All results are reported in Table~\ref{tab:SOTA_v2}.

\noindent{\textbf{Results.}} Compared with all existing DA methods, KDDAug-ECL achieves the best OOD and trade-off performance on two datasets. For UpDn, KDDAug-ECL improves the OOD performance of UpDn with a 21\% absolute performance gain and improves accuracies on all different question categories. For LMH, KDDAug-ECL boosts the performance on both benchmarks. 

\subsection{Ablation Studies}
\noindent{\textbf{Settings.}} To validate the effectiveness of gradually removing less-valuable samples in ECL, we compared different versions of ECL. To distinguish, we denote the version of ECL that does not remove less-valuable samples as \textbf{CL}, \ie, $q = 0$ for \textbf{CL}. We compared \textbf{ECL} and \textbf{CL} on four VQA backbones, and all results are shown in Table~\ref{tab:ECL}.

\noindent\textbf{Results.} From the results, we can observe that almost all VQA models can benefit from gradually removing less-valuable samples, with both ID and OOD performance improved. Furthermore, ECL can consistently improve the accuracy of \texttt{Other} in most settings, where the performance of \texttt{Other} is more reliable for evaluation~\cite{teney2020value}, which demonstrates the effectiveness of our ECL.

\section{Conclusions}

In this paper, we proposed a novel ECL to train VQA models on augmented samples effectively. ECL progressively trains VQA models from easy samples to hard samples, meanwhile dynamically removing less valuable ones. We validated the effectiveness of ECL on different DA methods through extensive experiments. Moving forward, we are going to extend our ECL to more general cases, such as DA methods in other VL tasks.

\noindent\textbf{Ethics Statement.}
Since ECL trains models based on augmented data, it may face the same potential ethical concerns as other existing DA works, such as training with augmented samples with inappropriate content. Apart from these general issues that already exist in the DA, our paper has no additional ethical issues.

\noindent\textbf{Limitations.}
While ECL calculated the difficulty scores during training iterations based on the current VQA model, the precise tuning of difficulty scores and their generalizability across various VQA architectures and datasets are open questions, suggesting areas for refinement in future research.

\bibliography{anthology,custom}

\appendix

\section*{Appendix}
\label{sec:appendix}
The Appendix is organized as follows:
\begin{enumerate}[leftmargin=1cm]

    \item[$\bullet$]  In Sec.~\ref{sec:implement}, we show more details about the experiment settings.
    
    \item[$\bullet$]  In Sec.~\ref{sec:para}, we show the hyperparameters analysis of ECL.

\end{enumerate}

\section{Details for Experiment Settings}
\label{sec:implement}
\textbf{Evaluation Datasets.} We evaluated the proposed KDDAug-ECL on two datasets: the ID benchmark \textbf{VQA v2}~\cite{goyal2017making} and OOD benchmark \textbf{VQA-CP v2}~\cite{agrawal2018don}. VQA v2 is a ``balanced" VQA dataset, where each question has complementary images with opposite answers. Although VQA v2 has reduced language biases to some extent, the statistical biases from questions still can be leveraged~\cite{agrawal2018don}. To disentangle the biases and clearly monitor the progress of VQA, VQA-CP re-organizes VQA v2, and deliberately keeps different QA distributions in the training and test sets. 

\textbf{Evaluation Metrics.} For model accuracies on VQA v2 and VQA-CP v2, we followed standard VQA evaluation metric~\cite{antol2015vqa}, and reported accuracy on three different categories separately: Yes/No (\texttt{Y/N}), number counting (\texttt{Num}), and other (\texttt{Other}) categories. For the ID evaluation, we reported the results on the VQA v2 val set. For the OOD evaluation, we reported the results on the VQA-CP v2 test set. Meanwhile, we followed~\cite{niu2021counterfactual} and used Harmonic Mean~(\textbf{HM}) of the accuracies on both two datasets (VQA v2 val \& VQA-CP test) to evaluate the trade-off between ID and OOD evaluations. 

\textbf{VQA Models.} Since KDDAug is an architecture-agnostic DA method, following KDDAug, we evaluated the effectiveness of our KDDAug-ECL on multiple different VQA models: UpDn~\cite{anderson2018bottom}, LMH~\cite{clark2019don}, RUBi~\cite{cadene2019rubi} and CSS~\cite{chen2020counterfactual,chen2021counterfactual}. Specifically, UpDn is a simple but effective VQA model, which always serves as a backbone for other advanced VQA models. LMH, RUBi, and CSS are SOTA ensemble-based VQA models for debiasing. For each specific VQA baseline, we followed their respective configurations (\eg, hyperparameter settings) and re-implemented them using the official codes. 

\textbf{Model Training.} We first pre-trained VQA models with only original samples following their respective settings. Then, we applied our ECL strategy and fine-tuned pre-trained VQA models with augmented samples from KDDAug for 5 epochs. Specifically, We only fine-tune the basic VQA backbone (UpDn) in the fine-tuning stage, \ie, for ensemble-based models, we removed the auxiliary question-only branches. For ECL, we set pace parameters $p=0.10$ and $q=0.10$. The batch size was set to 512. We used the Adamax~\cite{diederik2015adam} as the optimizer and the random seed was set to 0.

\section{Hyperparameters Analysis}
\label{sec:para}

\begin{figure}[t]
    \centering
    \subfloat{
        \includegraphics[width=\linewidth]{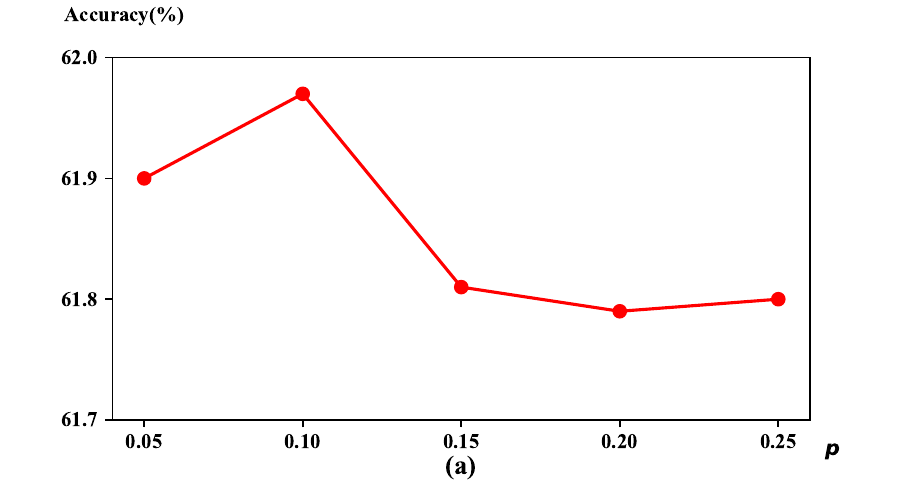}
    } \\
    \subfloat{
        \includegraphics[width=\linewidth]{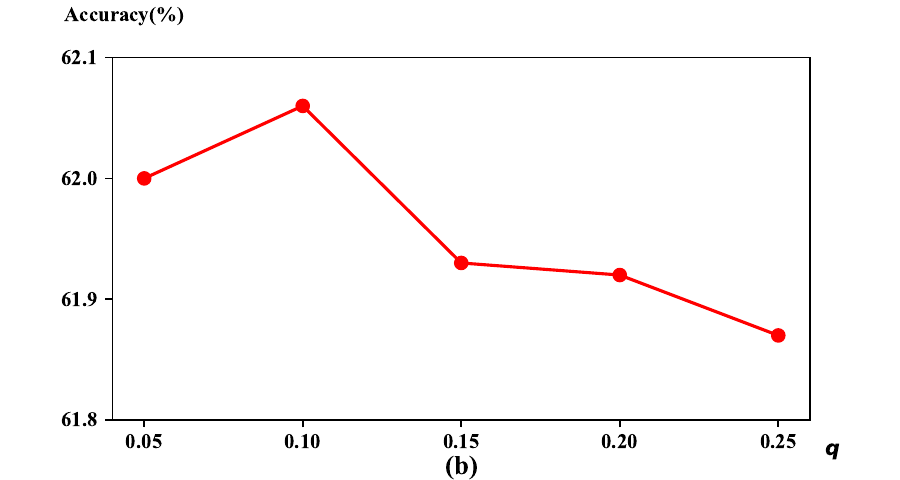}
    }
    \caption{Accuracies~(\%) on VQA-CP v2 test set of different hyperparameters settings of KDDAug-ECL. (a) The results of different values of pace parameter $p$. All results come from model CSS$^+$+KDDAug-ECL and $q$ is set to $0$. (b) The results of different values of pace parameter $q$. All results come from model CSS$^+$+KDDAug-ECL and $p$ is set to $0.10$.}
    \label{fig:influece_pq}
\end{figure}

We run a number of experiments to analyze the influence of hyperparameters $p$ and $q$. In order to make the curriculum learning progress at a suitable pace, we initially control the hyperparameters $p$ and $q$ in the range $\{0.05, 0.10, 0.15, 0.20, 0.25\}$. To rule out the effect of the step of removing less-valuable samples, we set $q$ to 0 when exploring the influence of $p$, \ie, not removing any samples. In the subsequent exploration of the influence of $q$, we set $p$ to $0.10$ since KDDAug-ECL achieved the best effect under this $p$ value. All results are reported in Figure~\ref{fig:influece_pq}. From the results, we can observe that the performance is best when $p=q=0.10$. In all the following experiments, we used the same best hyperparameter settings for KDDAug-ECL with different backbones.

\end{document}